\documentclass[10pt,journal,compsoc]{IEEEtran}

\usepackage{lineno}
\usepackage{graphicx}
\usepackage{amsmath}
\usepackage{amssymb}
\usepackage{algorithmic}
\usepackage[ruled]{algorithm2e}
\usepackage{array}
\usepackage{booktabs}
\usepackage{multirow}
\usepackage{stfloats}
\usepackage[pagebackref=true,breaklinks=true,colorlinks,bookmarks=false,citecolor=red]{hyperref}

\begin{document}

\title{Adaptive Scene Category Discovery with Generative Learning and Compositional Sampling}
%
%
%

\author{Liang~Lin,~\IEEEmembership{Member,~IEEE,}
        Ruimao~Zhang,~
        and~Xiaohua~Duan,
\IEEEcompsocitemizethanks{\IEEEcompsocthanksitem L. Lin is with the Key Laboratory of Machine Intelligence and Advanced Computing (Sun Yat-Sen University), Ministry of Education, China, with the School of Advanced Computing, Sun Yat-Sen University, Guangzhou 510006, P. R. China, and with the SYSU-CMU Shunde International Joint Research Institute, Shunde, China.\protect\\
E-mail: linliang@ieee.org
\IEEEcompsocthanksitem R. Zhang and X. Duan are with Sun Yat-Sen University, Guangzhou 510006, P. R. China.}
\thanks{This work was supported by the Hi-Tech Research and Development (863) Program of China (no. 2013AA013801), Guangdong Science and Technology Program (no. 2012B031500006), Guangdong Natural Science Foundation (no. S2013050014548), Special Project on Integration of Industry, Education and Research of Guangdong Province (no. 2012B091100148), and Fundamental Research Funds for the Central Universities (no. 13lgjc26).\protect\\Copyright (c) 2014 IEEE. Personal use of this material is permitted.
However, permission to use this material for any other purposes must be obtained from the IEEE by sending an email to pubs-permissions@ieee.org.}}


\markboth{IEEE Transactions on Circuits and Systems for Video Technology, 2014. }%
{L. Lin, \MakeLowercase{\textit{et al.}}: Adaptive Scene Category Discovery with Generative Learning and Compositional Sampling}
%


\maketitle

\begin{abstract}
This paper investigates a general framework to discover categories of unlabeled scene images according to their appearances (i.e., textures and structures). We jointly solve the two coupled tasks in an unsupervised manner: (i) classifying images without pre-determining the number of categories, and (ii) pursuing generative model for each category. In our method, each image is represented by two types of image descriptors that are effective to capture image appearances from different aspects. By treating each image as a graph vertex, we build up an graph, and pose the image categorization as a graph partition process. Specifically, a partitioned sub-graph can be regarded as a category of scenes, and we define the probabilistic model of graph partition by accumulating the generative models of all separated categories. For efficient inference with the graph, we employ a stochastic cluster sampling algorithm, which is designed based on the Metropolis-Hasting mechanism. During the iterations of inference, the model of each category is analytically updated by a generative learning algorithm. In the experiments, our approach is validated on several challenging databases, and it outperforms other popular state-of-the-art methods. The implementation details and empirical analysis are presented as well.
\end{abstract}




\section{Introduction}

Category discovery for unlabeled images is an important research topic with a wide range of applications such as content-based image retrieval~\cite{CoherentPhraseModel,NDRetrieval}, image database management~\cite{ImageHierarch,Zheng:MM04}, and scene understanding~\cite{I2T,GIST,Tuytelaars:IJCV09}. In this paper, we develop a unified framework to categorize scene images in an unsupervised manner. Specifically, with this framework, a batch of unlabeled scene images can be automatically grouped into different categories according to their contents, and we simultaneously generate the probability models for the categories.

We pose the unsupervised image categorization as a graph partition task, i.e., each generated partition indicates a potential category; then we employ a novel clustering sampling algorithm for inference, which is an extension of Swendsen-Wang cuts~\cite{swc} for greatly improving the inference efficiency. More specifically, the graph partition is formulated under a probabilistic framework that accumulates the generative models of all categories. Intuitively, the goodness of partitions is determined based on how well the learned models explain or generate the partitioned categories. Therefore, solving the optimal graph partition is equivalent to searching the maximum probability.

Natural scenes usually contain diverse image contents related with different types of visual appearance patterns, e.g., inhomogeneous (or structural) textures (buildings, cars, roads, etc.), and homogeneous textures (grasses, water surfaces, etc.)~\cite{LinGrammar}. Many studies~\cite{Mikolajczyk:PAMI05,Carneiro:CVPR03,Moreels:ICCV05} on designing image features show that the distribution-based descriptors (e.g., SIFT~\cite{sift}, HOG~\cite{HOG} and Textons~\cite{Textons}) and the binary operators (e.g., LBP and its variants~\cite{lbp,cs-lbp}) lead to state-of-arts on representing low-level image contents from different aspects. The former features tend to well describe the inhomogeneous textures, while the latter can be applied to capture highly random textures~\cite{LinPatch}. Therefore, in our method, we represent an image with a number of image patches at multiple scales. Two effective image features, the Histogram of oriented gradients (HOG)~\cite{HOG} and the Center-Symmetric Local Binary Pattern (CS-LBP)~\cite{cs-lbp}, are employed to describe the image patches. Specifically, we define two types of visual words (i.e., inhomogeneous textural words and homogeneous textural words), respectively, based on the two features. In literature, the significance of using combined features is also demonstrated in various vision tasks, e.g., near-duplicate image retrieval~\cite{CoherentPhraseModel,NDRetrieval}, object detection~\cite{HoGLBP}, and video tracking~\cite{LiuTracking,LinTrajectory}.

\begin{figure}
\centering
\includegraphics[width=3.5 in]{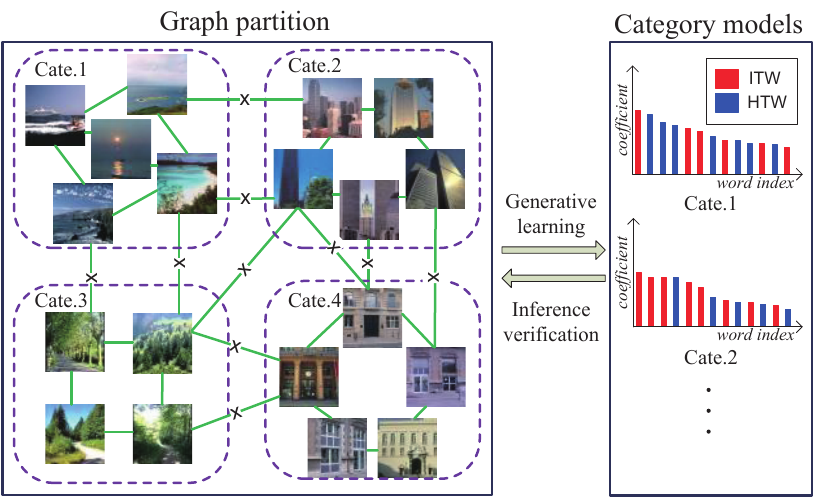}
\caption{An overview of our framework. We formulate the problem of image category discovery as a graph partition task. In the left panel, the images are treated as graph vertices that are partitioned into subgraphs by turning off the graph edges. As shown in the right panel, the generative models for all partitioned categories are pursued simultaneously, and the models are also used to guide the inference of graph partition. The models are learned with two types of visual words: inhomogeneous textural words (ITWs) and homogeneous textural words (HTWs) defined based on two image descriptors.}
\label{fig:framework}
\end{figure}

Moreover, we adaptively select informative features (i.e., visual words) for each scene class, along with the categorization procedure. Several methods of image categorization\cite{Sheikholeslami:MM98,DaiCVPR10} show that different categories of images are probably captured by different class-specific features. Some discriminative learning algorithms (e.g., Adaboost~\cite{Adaboost} and SVM) perform very well in feature selection. However, they are not suitable for our task, since these algorithms rely on negative data and are often sensitive to outliers. In contrast, our framework employs a generative learning algorithm based on information criteria~\cite{inducingfeatures,activebasis}, so that we can fast pursue the generative models of categories without extra negative data.

The framework of our approach is illustrated in Fig.\ref{fig:framework}. The key contribution of this work is a general approach for automatic scene image categorization, in which the cluster (i.e., category) number is automatically determined. The generative category models are learned and updated simultaneously together with the categorization procedure. Our method is evaluated on several public datasets and outperforms the state-of-the-art approaches. It is worth mentioning that the graph partition and category models are closely coupled. Given a state of partition, we can learn (or update) the probability models while the category models can drive the partition to be refined.

\subsection{Related Work}

Most of the methods of scene image categorization involve a procedure of supervised learning, i.e., training a multi-class predictor (classifier) with the manually labeled images~\cite{Jeon:MM04}. Unsupervised image categorization is often posed as clustering images into groups according to their contents (i.e., appearances and/or structures).  In some traditional methods\cite{Sheikholeslami:MM98}, various low-level features (such as color, filter banks, and textons~\cite{Textons}) are first extracted from images, and a clustering algorithm (e.g., $k$-means or spectral clustering) is then applied to discover categories of the samples.  

To handle diverse image content, some effective image representations such as bag-of-words (BoWs) are proposed~\cite{LiFeifei:CVPR05,Sivic:objectcate}, and they represent an image by using a pre-trained collection (i.e., dictionary) of visual words. Furthermore, Lazebnik et al.~\cite{SPM} present a spatial pyramid representation of BoWs by pooling words at different image scales, and this representation effectively improves results for scene categorization~ \cite{BogOfTexton}. Farinella et al.~\cite{ConstrainedDomain} propose to build an effective scene representation based on constrained and compressed domains.

To exploit the latent semantic information of scene categories, Bosch et al.~\cite{Bosch06} discuss the probabilistic Latent Semantic Analysis (pLSA) model that can explain the distribution of features in the image as a mixture of a few ``semantic topics''.  As an alternative model for capturing latent semantics, the Latent Dirichlet Allocation (LDA) model~\cite{Blei:LDA} was widely used as well. 

On the other hand, the category number is required to be predetermined or be exhaustively selected in many previous unsupervised categorization approaches~\cite{Tuytelaars:IJCV09,Liu:iccv2007}. In computer vision, the stochastic sampling algorithms~\cite{swc,LinGraphMatch,LinTrajectory} are shown to be capable of flexibly generating new clusters, merging and removing existing clusters in a graph representation. Motivated by these works, we propose to automatically determine the number of image categories with the stochastic sampling.

The rest of this paper is organized as follows. We first introduce the image representation in Section \ref{sec:imagerepresentation}.
Then we present the problem formulation in Section \ref{sec:formulation}, and follow with a description of the inference algorithm for unsupervised image categorization in Section \ref{sec:SWC}. Section \ref{sec:modellearning} discusses the learning algorithm for category model pursuit during the inference procedure. The experimental results and comparisons are exhibited in Section \ref{sec:experiments}, and the paper is concluded in Section \ref{sec:conclusion}.


\section{Image Representation}
\label{sec:imagerepresentation}

In this section, we start by briefly introducing the two effective low-level image descriptors used in this work, and define two types of visual words to construct the dictionary of images.




%
%
%
%

%

%

\begin{figure}
\centering
\includegraphics[width=3.4 in]{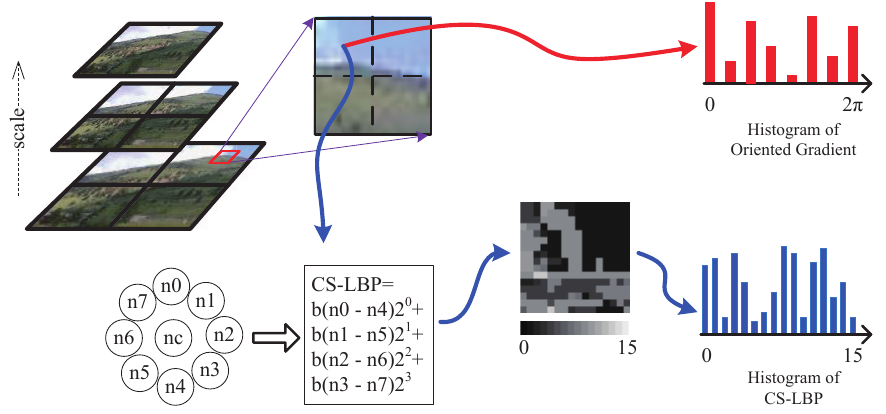}
\caption{Image representation. We represent an image with the pyramid Bag-of-Words (BOW) model with two types of visual words that are, respectively, defined based on  two image descriptors, i.e., HOG~\cite{HOG} and CS-LBP~\cite{cs-lbp}.}
\label{fig:image-representation}
\end{figure}

Previous works on designing image features can be roughly divided into two categories~\cite{LinPatch,HoGLBP}. The first one explicitly describes images with local gradients that are sensitive to structures (e.g., edges, boundaries, and junctions) and distinct textures (e.g., regions of clear details). The other one reflects uncertain differences among pixels and thus tends to be suitable for incognizable random textures (e.g., complex regions, and cluttered patterns). Thus, we utilize two typical image descriptors, i.e., HOG~\cite{HOG} and CS-LBP~\cite{cs-lbp}, respectively, in this work. Following the studies on image representation~\cite{LinPatch}, we refer a visual word $\omega$ as an ensemble or equivalence class of image patches that share the similar appearances. Letting $h(\cdot)$ be the histogram of an image feature, we define $\omega$ as,

\begin{equation}\label{eq:visualword}
\omega=\big\{\Lambda:h(\Lambda)=\hat{h}+\epsilon\big\},
\end{equation}
where $\hat{h}$ denotes the mean histogram of the image patches, and $\epsilon$ is the statistical fluctuation, i.e., a very small value. According to the two image descriptors, we define two types of visual words, inhomogeneous textural words (ITWs) and homogeneous textural words (HTWs), together with the two descriptors. The benefit of combining the two types of words will be demonstrated in the experiments.

To define ITWs, the input image domain is divided into a number of regular cells; at each pixel, a local gradient is calculated, and a histogram is pooled over each cell for different orientations. As illustrated in Fig.~\ref{fig:image-representation}, we decompose an image patch by $2 \times 2$ cells and quantize the orientations into $8$ angles. The dimension of this descriptor is thus $32$.

The HTWs are generated using the CS-LBP operator, which is computed at every pixel in the input image domain. It compares center-symmetric pairs of the given pixel  and forms a binary vector. Given a pixel located at $x$ with $\hat{n}=8$ neighborhood pixels that are equally spaced on a circle of radius, as the example illustrated in Fig.~\ref{fig:image-representation}, the binary vector can be calculated as,

\begin{equation}
\label{eq:cs-lbp}
\begin{split}
\sum_{i=0}^{\hat{n}/2-1}b(n_i-n_{i+\hat{n}/2})2^i,\;\;\;\;\; b(x)=\begin{cases}1, & x>1\\0, & otherwise \end{cases}
\end{split}
\end{equation}
where $n_i$ and $n_{i+\hat{n}/2}$ correspond to the intensity scales of center-symmetric pairs of pixels. We compute the operator over all pixels in the domain; the obtained binary vectors can be converted into decimal strengths in the range of $[0,15]$. An example of a strength map is shown in Fig.\ref{fig:image-representation}. Since there are 4 cells divided, we further pool the strengths into a histogram with $16 \times 4 = 64$ bins, denoted as $h^b$.




Then we construct the dictionary to represent images with the visual words. In our implementation, we collect a large number of image patches from our database and compute the two descriptors for each, and group them into a batch of clusters (words) using the k-means algorithm. Thus, we obtain a dictionary $\mathcal{W}=\{\omega_i, i = 1, \ldots, m\}$, where $\omega_i$ is a visual word (i.e. ITW or HTW).

Given an image $\mathbf{I}$, we represent it with a spatial pyramid format, $ 1 + 4 \times 2 = 9$ blocks, i.e., $3$ scales (resolutions) and $4$ blocks in each scale except the top, as illustrated in Fig.~\ref{fig:image-representation} . In each block, the image domain is further decomposed into regular image patches that are mapped to the generated words. The image of a block $\mathbf{J}$ can be thus represented as a vector by using the dictionary, $(r_1(\mathbf{J}), r_2(\mathbf{J}), \dots, r_m(\mathbf{J}))$, where $r_i(\mathbf{J})$ is the response with the visual word $\omega_i$, and

\begin{equation}
\label{eq:word_response}
r_i(\mathbf{J})=\psi\bigg(\sum_{\Lambda\in{\mathbf{J}}} {\bf{1}}_{\omega_i}(\Lambda) \bigg),
\end{equation}
where ${\bf{1}}_{\omega_i}(\Lambda) = \{ 1 | 0 \}$, the indicator function, is used to indicate whether the image patch $\Lambda \in \mathbf{J}$ matches with  $\omega_i$.  The matching is measured by either of the two descriptors, $h^a$ and $h^b$, according to the type of word $\omega_i$. Thus, we use  $\sum_{\Lambda\in{\mathbf{J}}} {\bf{1}}_{\omega_i}(\Lambda)$ to indicate the number of the visual word $\omega_i$ matching with the image block $\mathbf{J}$. Here $\psi(\cdot)$ is the sigmoid function $\delta(\cdot)$ that is characterized by a saturation level.
%

The image $\mathbf{I}$ is hence represented as $\mathcal{R}(\mathbf{I})$, by concatenating the vectors of all $9$ blocks.

\section{Problem Formulation}
\label{sec:formulation}

Given a set of unlabeled images $\mathcal{D}$, the goal of our framework is to categorize them into an unknown number of disjoint $K$ clusters, as

\begin{equation}
\label{eq:partition}
\Pi \!\!= \!\!\{\pi_1, \pi_2, \dots, \pi_K\},
\end{equation}
where $\cup_{k=1}^K{\pi_k}= \mathcal{D}, \pi_i\cap \pi_j=\emptyset,\;\forall i\neq j$.

We first build a graph $G_0=\langle V, E_0\rangle$, in which $V = \mathcal{D} =\{\mathbf{I}_1, \mathbf{I}_2, \dots, \mathbf{I}_N\}$ is the set of graph vertices specifying the images to be categorized, and $E_0$ is the set of edges connecting neighboring graph vertices. Then we solve the task of graph partition by cutting edges of the graph, i.e., generating disjoint subgraphs. 
However, $G_0$ is a fully connected graph where the initial edge set $E_0$ could be very large. To reduce computational complexity, we shall compute a relatively sparse graph representation $G_0=\langle V, E\rangle$ by pruning edges, $ E \subset E_0$.

For any edge $e \in E_0$, an auxiliary connecting variable $\mu_e = \{ \text{on} | \text{off} \}$ is first introduced, which indicates whether the edge is turned on or off. Then we can define the edge connecting probability by measuring the similarity of two connected graph vertices. In our implementation, We define the similarity using the visual words $\mathcal{W}$. Specifically, for any vertices $v \in V$, we represent it as, $\mathcal{R}(\mathbf{I}) = ( r_1(\mathbf{I}), r_2(\mathbf{I}), \ldots, r_m(\mathbf{I}))$, where $r_i(\mathbf{I})$ is the response of the word $\omega_i$, as in Equation~(\ref{eq:word_response}). Thus, we can define the connecting probability $q_e$ for two arbitrary images $\mathbf{I}_s \in V, \mathbf{I}_t \in V$ as,


\begin{equation}
\label{eq:edgeprob}
q_e(s,t)= p (\mu_e = on | v_s, v_t) = \exp\bigg\{-\tau\big{[\mathcal{KL}(\mathcal{R}_s\| \mathcal{R}_t ) \big]}\bigg\},
\end{equation}
where we denote $\mathcal{R}_s = \mathcal{R}(\mathbf{I}_s)$ and $\mathcal{R}_t = \mathcal{R}(\mathbf{I}_t)$ for notation simplicity. $\mathcal{KL}()$ is the symmetric Kullback-Leibler distance for measuring two feature vectors. $\tau$ is a constant parameter. $q_e(s,t)$ should be close to $0$ if $\mathbf{I}_s$ and $\mathbf{I}_t$ naturally belong to different categories; the edge $e$ connecting $\mathbf{I}_s$ and $\mathbf{I}_t$ could be then turned off with high probability. 

In practice, the edges with very low turn-on probability can be directly removed. Furthermore, we enforce each vertex can be only connected to at most $6$ neighbors. That is, for any vertex we keep $6$ edges with the highest connecting probabilities, and remove the other edges. Therefore, we obtain the sparse graph $G= \langle V, E\rangle$ where $E \subset E_0$.

With the graph representation, we pursue the generative probability models for all categories, as

\begin{equation}
\Phi =  \{\phi_k(\mathbf{I}; W_k, \Theta_k), W_k \subset \mathcal{W}, k=1,\dots,K\} ,
\end{equation}
where $W_k \subset \mathcal{W}$ denotes the selected visual words for modeling the category $\pi_k$ and $\Theta_k$ includes the corresponding model parameters, i.e., the coefficients of words. The overall solution of image category discovery can be defined as,

\begin{equation}
\label{eq:solution}
S=\big(K, \Pi, \Phi),
\end{equation}
where $K$ is the inferred category number. The graph partition $\Pi$ and category modeling $\Phi$ can be solved together in a Bayesian inference framework. Assume that $p(S)$ and $p( \mathcal{D} | S)$ denote the prior model and the likelihood model, respectively. $p(S)$  can be simply modeled by incorporating an exponential function for $K$, as we impose no priors on $\Pi$ and $\Phi$. The likelihood model $p( \mathcal{D} | S ) = p(\mathcal{D} | \Pi , \Phi)$ can be defined as a product of generative models of all separated categories, as we assume the models are generated independently to each other. We can then define the posterior probability of solution $S$ as,

\begin{equation}\label{eq:postp}
\begin{split}
    p(S | \mathcal{D}) &\propto p(S) p( \mathcal{D} | S) \\
    & =\exp\{-\beta K\} \prod_{k=1}^K \phi_k(\mathbf{I}; W_k, \Theta_k),
\end{split}
\end{equation}
where $\beta$ is an empirical parameter for constraining the number of inferred categories. The category model $\phi_k(W_k, \Theta_k)$ is defined on the probabilistic distribution of the images in partition $\pi_k$. The models for all categories can be learned and updated during the procedure of image categorization.


%
%

\section{Inference for Image Categorization}
\label{sec:SWC}

The objective of inference is to search for the optimized solution $S^*$ by maximizing the posterior probability in Equation~(\ref{eq:postp}),

\begin{equation}
S^* = \arg\max p(S | \mathcal{D} ).
\end{equation}
This optimization is very challenging due to two characters in our problem: (i) the unknown number of partitions, (ii) no confident initializations, i.e., lack of the initial category models. Therefore, we employ the stochastic sampling algorithm instead of using deterministic inference algorithms.


%
%

In the research area of stochastic inference, cluster sampling is very powerful for simulating Ising/Potts graphical models, which is designed under the Metropolis-Hasting mechanism. Recently, Barbu and Zhu~\cite{swc} generalized  the algorithm, namely Swendsen-Wang cuts (SWC), to solve graph partition in several vision applications. This algorithm enables us to effectively search for the maximum of posterior probability. It simulates a Markov chain containing a sequence of states in the solution space $\Omega$ and visits the Markov chain by realizing a reversible jump between any two successive states.

\begin{figure*}[!htbp]
\centering
\includegraphics[scale=0.85]{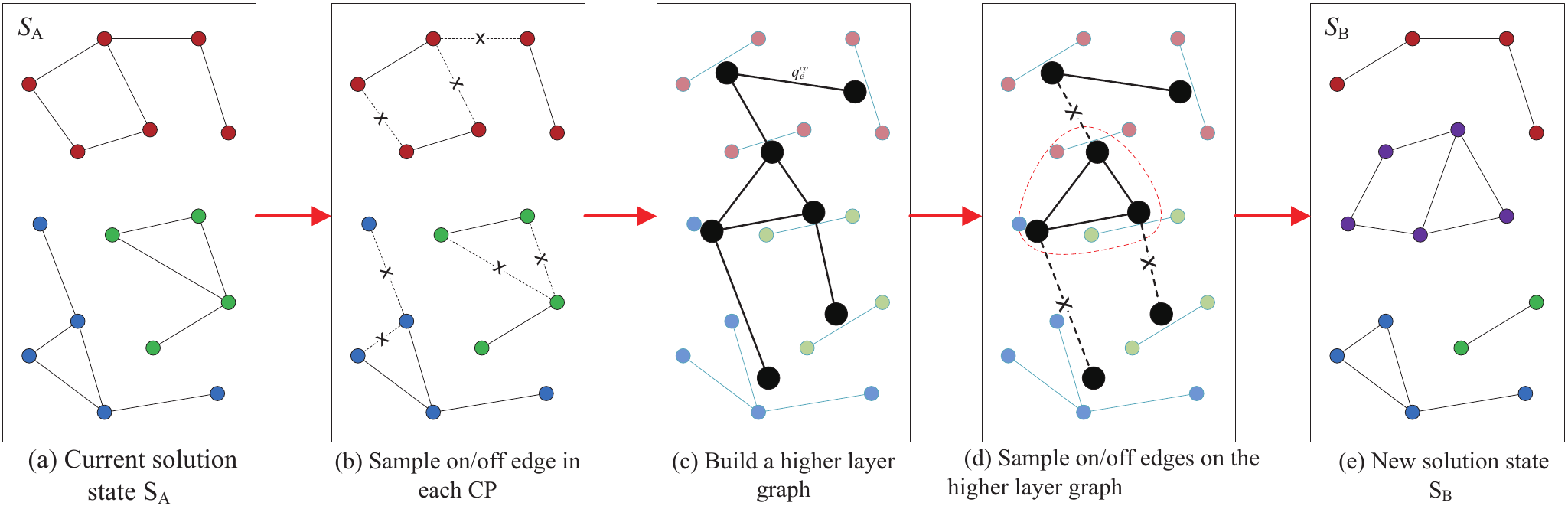}
\caption{Illustration of the compositional Swendsen-Wang cuts algorithm for exploring a new solution state.}
\label{fig:mwayswc}
\end{figure*}

In the following, we first introduce the SWC algorithm, and then discuss an extension~\cite{DuanVideo} that greatly improves the inference efficiency. In general, the SWC algorithm iterates in two steps:

\begin{enumerate}
    \item Generate the connected components ($CP$s) by probabilistically turning off connecting edges in the graph. Graph vertices connected together by ``on'' edges form a connected component (denoted by $CP$ for simplicity). Specifically, any two vertices in one $CP$ are linked by a path that consists of several edges. For arbitrary edge $e \in E$, we sample its connecting variable $\mu_e$ and decide it is turned on or off in this step. Then we obtain a few $CP$s, each of which is a set of connected graph vertices.
    \item Explore a new partition solution by relabeling one of the $CP$s. Assume that the current partition solution is $S_A$ and we are exploring a new solution $S_B$. Given one randomly selected $CP$, the reversible operators are developed to re-assign its label. For example, the selected $CP$ can be merged into current separated category by receiving the same label with the category; otherwise, a new category can be created if the selected $CP$ receives a new label.
\end{enumerate}

We design the algorithm by the Metropolis-Hastings mechanism~\cite{swc}. Let $Q(S_A \rightarrow S_B)$ be the proposal probability for moving from state $S_A$ to state $S_B$, and conversely, $Q(S_B \rightarrow S_A)$ is the proposal probability from $S_B$ to $S_A$. The acceptance rate of the moving from $S_A$ to $S_B$ is,

\begin{equation}
\label{eq:acceptprob}
\alpha(S_A\rightarrow S_B)=\min\bigg(1,\frac{Q(S_B\rightarrow S_A)}{Q(S_A\rightarrow S_B)}\cdot\frac{p(S_B|\mathcal{D})}{p(S_A|\mathcal{D})}\bigg).
\end{equation}
For any state transition, the proposal probability usually involves two aspects: (i) the generation of $CP$, and (ii) the label assignment of $CP$. In our method, we make the $CP$ be assigned randomly with a uniform distribution, so that the proposal probability can be simplified.  Thus, the ratio of proposal probability is calculated by,\begin{equation}
\frac{Q(S_B\rightarrow S_A)}{Q(S_A\rightarrow S_B)} = \frac{\prod_{e \in C_B} (1 - q_e)}{ \prod_{e \in C_A} (1 - q_e) },
\end{equation}
where $C_A$ denotes the edge set of edges that are probabilistically turned off for generating the $CP$ on state $S_A$, and similarly $C_B$ is the turning-off edge set on $S_B$. Here we name $C_A$ or $C_B$ as a ``cut'', following \cite{swc}.


To further accelerate the convergence of inference, we employ an improved version of the SWC algorithm that was originally proposed by us for video shot categorization~\cite{DuanVideo}.  In the original algorithm, only one $CP$ is selected and processed in each step of solution exploration. In our method, we process a number of $CP$s together by coupling them into a combinatorial cluster. We thus regard this algorithm as the compositional SWC (CSWC). The CSWC algorithm is able to enlarge the searching scope during the sampling iterations, resulting in faster convergence than the original version.  

Fig.~\ref{fig:mwayswc} illustrates the idea of CSWC.  Given a current state $S_A$ ( as shown in Fig.\ref{fig:mwayswc} (a)), we can generate a number of $CP$s by turning off a few edges (as shown in Fig.\ref{fig:mwayswc} (b)). Then we construct a higher layer graph  $\mathbf{G}$ based on these $CP$s. In this graph, we treat each $CP$ as a vertex, and link any two neighboring $CP$s by an edge, as shown in Fig.\ref{fig:mwayswc}(c). Within $\mathbf{G}$, we can generate the combinatorial cluster, where several $CP$s are selected.

Similar with the definitions in $G$, we calculate the turn-on probability $q^{CP}$ for an edge in $\mathbf{G}$ according to the similarity of two connected vertices (i.e., $CP$s), which can be derived from the original graph $G$. Specifically, given two neighboring $CP_i$ and $CP_j$, we measure their similarity by aggregating all the edges in $G$ that connects the vertices in $G$ belonging to $CP_i$ and $CP_j$, respectively. Thus, we define the edge probability in $\mathbf{G}$ as,

\begin{eqnarray}
 q^{CP} &\propto& \big[ 1 - \prod (1 - q_e)  \big], \\\nonumber 
&& e=<s,t>, s \in CP_i, t \in CP_j.
\end{eqnarray}

By probabilistically turning off the edges in $\mathbf{G}$, we can also generate several connected components, and we regard them as combinatorial clusters to distinguish the $CP$s in $G$. In Fig.\ref{fig:mwayswc}(d), $4$ combinatorial clusters are generated. Different with the algorithm in \cite{DuanVideo}, we allow more than one combinatorial clusters to be selected in this step, and we assign labels to the them. In this way, we generate a new solution of graph partition accordingly. In the implementation,  we enforce each combinatorial cluster being processed as a atomic unit, i.e., all original $CP$s in the compositional cluster will receive the same label.  As Fig.\ref{fig:mwayswc} illustrates, to go from $S_A$ to $S_B$, the original SWC algorithm needs at least three steps, whereas for CSWC there is only one step. Note that we visualize only one selected CP in Fig.\ref{fig:mwayswc} (d) for illustration.

During the inference, the posterior probability $p(S | \mathcal{D} )$ can be changed, as we keep the category models updated with the categorization operation. Note that we only need to update the models of the categories where we add or remove images within them. We will introduce the category model learning in the next section.



\begin{algorithm}[!ht]
\caption{The sketch of our approach}
\label{alg:shotcate}
\KwIn{Image dataset $\mathcal{D}=\{\mathbf{I}_1,\dots,\mathbf{I}_N\}$, and visual words $\mathcal{W}=\{\omega_1, \dots,\omega_M\}$}
\KwOut{The categorization solution $S=\big(K, \Pi, \Phi\big)$}

1. Initialization\;
~~~~(1) Represent each image $\mathbf{I}_i$ with the visual words, $\mathcal{R}(\mathbf{I}_i)=\{r_1(\mathbf{I}_i),\dots, r_m(\mathbf{I}_i)\}$. \\
~~~~(2) Create the graph $G_0=\langle V, E_0\rangle$, and compute the turn-on probability $q_e$ according to Equation~(\ref{eq:edgeprob}), $\forall e \in E_0$. \\
~~~~(3) Remove the edges with low turn-on probability deterministically, and generate the sparse graph $G=\langle V, E\rangle$. \\

2. Repeat for cluster sampling\;
~~~~(1) At the current solution $S_A$, generate the $CPs$ by probabilistically turning off connecting edges in the graph $G$.\\
~~~~(2) Construct a high layer of graph $\mathbf{G}$ based on $CPs$.\\
~~~~(3) Generate combinatorial clusters by probabilistically turning off edges in $\mathbf{G}$.\\
~~~~(4) Select several combinatorial clusters and re-assign labels to them.\\
~~~~(5) Accept the new solution $S_B$ according to the acceptance rate defined in Equation~(\ref{eq:acceptprob}).\\
~~~~(6) Update the generative models, $\phi(\mathbf{I}_{k,i}; W_k, \Theta_k)$, for the categories that have been modified according to solution $S_B$.\\
~~~~(7) Update the posterior probability $( S | \mathcal{D} )$ accordingly.\\

3. Output the final solution $S^* = \arg\max p( S | \mathcal{D} )$.
\end{algorithm}

\section{Category Model Learning}
\label{sec:modellearning}

Given a fixed graph partition $\Pi$, we learn the probability model $\phi_k(W_k, \Theta_k)$ for each category by selecting the most informative visual words. Since all scene images in $\mathcal{D}$ are unlabeled, and no extra negative samples are provided, we employ an efficient generative learning algorithm for this task, namely information pursuit~\cite{inducingfeatures,Zhu:HIT}. Similar approaches of combining generative learning in unsupervised categorization are discussed in \cite{DaiCVPR10}.

Suppose the category $\pi_k$ is governed by an underlying target model $\phi_{f,k}$, the model pursuit can be solved by additively searching for a sequence of features, starting from an initial model $\phi_{k,0}$. At each step $t$, the model $\phi_{k,t}$ is updated to gradually approach $\phi_{f,k}$. Here that we ignore $k$ for notation simplicity. In the manner of stepwise pursuit, the new model $\phi_t$ is updated by adding a new feature $\omega_t$ based on the current model $\phi_{t-1}$, and $\omega_t$ imposes an additive constraint, as,

%
%

\begin{eqnarray}
\phi_t = \frac{1}{z_t} \phi_{t-1} e^{\lambda_t r_t},\\\nonumber
 s.t. ~~
 E_{\phi_{t}} [ r_t ] =  E_{\phi_f} [ r_t ],
\end{eqnarray}
where $r_t$ denotes the response of the word $\omega_t$. $E_{\phi_f}(r_t)$ represents the expectation of feature $\omega_t$ over the underlying model, which can be calculated by averaging feature responses over positive samples. $E_{\phi_{t}} [ r_t ]$ denotes the feature expectation on the new model. Follow~\cite{inducingfeatures,Zhu:HIT}, we can derive the probability model by $T$ rounds of model pursuit as the following Gibbs form,

\begin{equation}\label{eq:ori_model}
\phi(\mathbf{I} ; W, \Theta)=\phi_0 (\mathbf{I}) \frac{1}{Z}\exp\bigg\{\sum_{t=1}^{T}\lambda_t r_t (\mathbf{I}) \bigg\},
\end{equation}
where $Z = \prod z_t$ and $\Theta=(\lambda_1,\dots,\lambda_T)$. $z_t$ normalizes the sum of the probability to $1$, and $\lambda_t$ is the coefficient weight of the selected feature $\omega_t$. In our implementation, we specify the initial model $\phi_0$ as a uniform distribution over all words.


With this definition in Equation~(\ref{eq:ori_model}), the model is updated by  solving $\lambda_t$ and $r_t$ at each round $t$. Here we discuss a MaxMin-KL algorithm for this goal, which iteratively performs with two following steps.

{\bf Step 1: Max-KL.} The most informative feature $r_t^*$ is selected to update the current model. This step optimizes the following problem, given the candidate features,

\begin{eqnarray}\label{eq:MaxKL}
r_t^* & = & \arg\max_{r_t} \mathcal{K}( \phi_t \| \phi_{t-1} ) \\\nonumber
      & = & \arg\max_{r_t} \lambda_t  E_{\phi_f} [ r_t ] - \log z_t.
\end{eqnarray}
This step could be computational expensive as we need to sample the model distribution $\phi_{t-1}$ of the previous round $t-1$. Following recent works on image template learning~\cite{Zhu:HIT,LinPatch}, we can simplify the computation by enforcing the visual words have little overlap. In particular, all features can be selected independently. The optimization in Equation~(\ref{eq:MaxKL}) can be approximated as,
\begin{equation}
r_t^* = \arg\max_{r_t}  E_{\phi_f} [ r_t ] -  E_{\phi_0} [ r_t ],
\end{equation}
where $E_{\phi_0} [ r_t ]$ can be ignored, as it is a constant calculated on the initial model $\phi_0$. We calculate $E_{\phi_f} [ r_t ]$ by the mean response values, 
\begin{equation}
E_{\phi_f} [ r_t ] = \frac{1}{n_k} \sum_{i=1}^{n_k} r_t(\mathbf{I}_i),
\end{equation}
where $n_k$ is the number of images belonging into the $k$-th category.

{\bf Step 2: Min-KL.} Given the selected feature $r_t$, this step is to compute its corresponding weight $\lambda_t$ and normalization term $z_t$ by 
\begin{eqnarray}\label{eq:MinKL}
\lambda_t^* = \arg\min_{\lambda_t} \mathcal{K}( \phi_t \| \phi_{t-1} ) \\\nonumber
      s.t. ~~
 E_{\phi_{t}} [ r_t ] =  E_{\phi_f} [ r_t ].
\end{eqnarray}
This optimization in Equation~(\ref{eq:MinKL}) can be solved analytically according to the proof in \cite{LinPatch}, and we conduct that,
\begin{eqnarray}
\lambda_t &=& \log \frac{ E_{\phi_f} [ r_t ] (1  -  E_{\phi_0} [ r_t ])}{  (1 - E_{\phi_f} [ r_t ] )E_{\phi_0} [ r_t ]} \\\nonumber
z_t &=& \exp{\lambda_t} E_{\phi_0} [ r_t ] + 1 - E_{\phi_0} [ r_t ].
\end{eqnarray}

Since we can analytically pursue this model by selecting a number $T$ of informative features, the model in Equation (\ref{eq:ori_model}) can be further simplified into the following form, 

\begin{equation}\label{eq:promodel}
\phi (\mathbf{I};\Theta)=\phi_0 (\mathbf{I}) \prod_t^T \bigg[ \frac{1}{z_t} \exp\{ \lambda_t r_t (\mathbf{I}) \}  \bigg].
\end{equation}

The proposed algorithm in the above is simple and fast, because the value of $E_{\phi_f} [r_t]$ and $E_{\phi_0} [ r_t ]$ for each feature only need to be computed once in the off-line stage. Hence, we can embed the learning algorithm to keep the category model updated during the iterating procedure of categorization.

%

Algorithm.\ref{alg:shotcate} summarizes the overall sketch of our framework.

\section{Experiments}
\label{sec:experiments}

In the experiments, we apply our method to discover categories for a batch of unlabeled images with diverse appearances, and compare with other state-of-the-art approaches. 


\begin{figure*}[htbp]
\includegraphics[width=7.0 in]{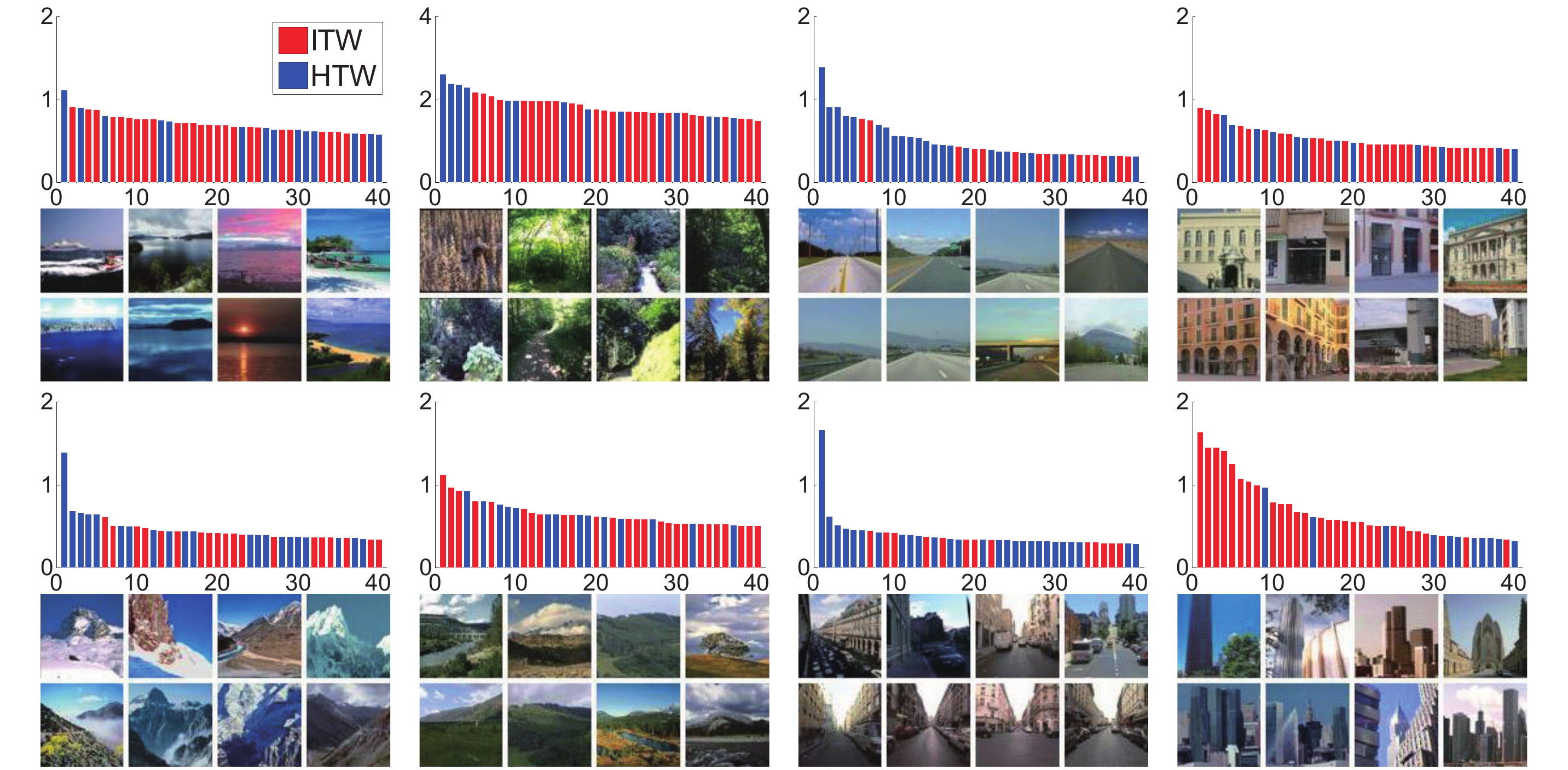}
\caption{The selected visual words for $15$ categories of the UIUC-Scene database. For each category, we show the top $40$ informative visual words according to their information gains (the vertical axis). The different colors represent different types of words (i.e., red for ITWs and blue for HTWs).}
\label{fig:ftrselct}
\end{figure*}

\subsection{Datasets and Metrics}

We use three challenging public databases for validation: MIT-Scene\footnote{http://people.csail.mit.edu/torralba/code/spatialenvelope/}, Corel\footnote{http://wang.ist.psu.edu/docs/related.shtml}, and UIUC-Scene\footnote{http://www-cvr.ai.uiuc.edu/ponce\_grp/data/index.html}. Moverover, these three databases are mixed together as a larger testing set for further evaluation. 

The MIT-Scene database contains $2688$ images classified into $8$ categories according to their meaningful semantics: coasts, forest, mountains, country, highways, city views, buildings, and streets. The number of images in each category is in the range of $260 \sim 410$, and the resolution of each image is $256 \times 256$ pixels. The Corel dataset includes $1000$ natural scenes with the resolution $256 \times 384$ pixels of 10 semantic categories: bus, coasts, dinosaurs, elephants, flower, food, horses, mountains, people, and temples. Each category contains $100$ images. The UIUC-scene database, which is an extension of MIT-Scene, contains $4485$ images classified into $15$ categories, and their themes are various, e.g., mountains, forest, offices, and living rooms. The mixed dataset is the union of all the three databases, including totally $5485$ images of $23$ categories. Note that there are a few overlapping categories among them.

\begin{table}[h]
\caption{The inferred cluster number in each time of experiment.}
\centering
\begin{tabular}{lcccccccccc}
\toprule
\# & 1 & 2 & 3 & 4 & 5 & 6 & 7 & 8 & 9 & 10 \\
\hline
I & 8  & 9 &  8 & 8  & 9  & 10 & 10 & 11 & 8 & 9 \\
\hline
II & 9 & 10 & 10 & 11 & 10 & 9 & 12 & 11 & 12 & 10 \\
\hline
III & 16  & 17 &  15 & 16  & 16  & 15 & 18 & 16 & 15 & 17 \\
\hline
IV & 27 & 24 & 24 & 26 & 24 & 25 & 24 & 26 & 25 & 25\\
\bottomrule
\multicolumn{11}{l}{\#: No. of experiments;}\\
\multicolumn{11}{l}{I: Experiments on the MIT database;}\\
\multicolumn{11}{l}{II: Experiments on the Corel database;}\\
\multicolumn{11}{l}{III: Experiments on the UIUC database;}\\
\multicolumn{11}{l}{IV: Experiments on the mixed dataset.}
\end{tabular}
\label{tab:clusternum}
\end{table}


\begin{table*}[bp]
\caption{Performance comparison via Purity (higher is better)}
\centering
\begin{tabular}{cccccccccc}
\toprule
 & \multirow{2}*{\small K-means} & \multirow{2}*{\small GIST}  & \multirow{2}*{pLSA} & \multirow{2}*{LDA} & \multirow{2}*{AP} & \multicolumn{3}{c}{Ours} \\
\cmidrule{7-9}
 &  &  &  & & & ITW+HTW & ITW & HTW\\
\midrule
MIT   & 0.5529 & 0.5770 &	0.6457	& 0.6096	& 0.5546 & \textbf{0.6721} &0.5764 & 0.6000 \\
\midrule
Corel & 0.5337 & 0.5644 &	0.6070	& 0.5980	& 0.5612 & \textbf{0.6203} & 0.6160 & 0.6040\\
\midrule
UIUC & 0.4487 & 0.4514 &	0.5074	& 0.5449 & 0.5850 & \textbf{0.5964} & 0.5613 & 0.5148\\
\midrule
Mixed & 0.3632 & 0.3801 &	0.4136	& 0.4801 & 0.5017 & \textbf{0.5295} & 0.4836 & 0.4226\\
\bottomrule
\end{tabular}
\label{tab:Purity}
\end{table*}

\begin{table*}[bp]
\caption{Performance comparison via Conditional Entropy (lower is better)}
\centering
\begin{tabular}{ccccccccc}
\toprule
& \multirow{2}*{\small K-means} & \multirow{2}*{\small GIST} & \multirow{2}*{pLSA} & \multirow{2}*{LDA} & \multirow{2}*{AP} & \multicolumn{3}{c}{Ours} \\
\cmidrule{7-9}
 &  &  &  &  &  & ITW+HTW & ITW & HTW\\
\midrule
MIT & 1.2465  & 1.2102&	1.0156	& 1.1836	& 1.1400 & \textbf{0.8963} & 1.1536& 1.1145 \\
\midrule
Corel  &  1.3105  & 1.2136 & 1.1234 &	1.1371 & 1.2577 &\textbf{1.0909} & 1.1036 & 1.1154\\
\midrule
UIUC &1.5020  & 1.4564 & 1.4322 & 1.3146 & 1.2121 &\textbf{1.1581} & 1.2150 & 1.3948\\
\midrule
Mixed &  1.7603  & 1.7172 & 1.6811 & 1.5828 & 1.5127 &\textbf{1.4328} & 1.4971 & 1.5955\\
\bottomrule
\end{tabular}
\label{tab:CondEnt}
\end{table*}

The usual evaluation metric for categorization is Average Precision, and the number of categories is assumed to be predetermined. In this work, we adopt the two recently proposed metrics for unsupervised categorization~\cite{Tuytelaars:IJCV09,DuanVideo}, i.e., {\em Purity} and {\em Conditional Entropy}. In brief, the larger value of \textit{Purity} implies the better performance in categorization and \textit{Conditional Entropy} inversely.

For the input set $\mathcal{D}$, including a number of $N$ images, suppose the underlying category number is $L$ and the corresponding groundtruth category labels are denoted by $X=\{x_i\in [1, L], i=1,\dots,N\}$. A testing system groups the images into $K$ categories, $\{D_k, k=1, \dots, K\}$, with the inferred category labels $Y=\{y_i\in [1, K], i=1, \dots, N\}$. It is worth mentioning that $K$ could be not equal to $L$, as we allow the algorithm to automatically determine the number of categories. The metric \textit{Purity} and \textit{Conditional Entropy} are defined as,

\begin{eqnarray}
Purity(X|Y)=\sum_{y\in Y}p(y)\max_{x\in X}p(x|y),\\
H(X|Y)=\sum_{y\in Y}p(y)\sum_{x\in X}p(x|y)\log{\frac{1}{p(x|y)}},
\end{eqnarray}
where $p(y)=\frac{|D_y|}{N}$ and $p(x|y)$ can be simply estimated from the observed frequencies in categorized data, resulting in an empirical estimation. $|D_y|$ represents the number of images in one category.

%

\subsection{Parameter settings and results}

We carry out the experiments on a PC with Quad-Core 3.6GHz CPU and 32GB memory.  We set the parameter $\beta = 300$ in the probabilistic formulation (in Equation~(\ref{eq:postp})), and the parameter $\tau=0.2$ in the probabilistic edge definition (in Equation~(\ref{eq:edgeprob})).

In our experiments, we first randomly collect a number of image patches with different scales from the datasets and generate $500$ ITWs and $500$ HTWs as introduced in Section~\ref{sec:imagerepresentation}. There are totally $1000$ words in the dictionary.

We carry out our method $10$ times and use the average performance for comparison. The inferred category number may not be identical each time, as reported in Table.\ref{tab:clusternum}. The average category number is $9.0$ for the MIT-Scene, $10.4$ for the Corel, $16.1$ for UIUC-Scene, and $25.0$ for the mixed dataset.

\begin{figure*}[!htbp]
\centering
\includegraphics[width=6.8 in]{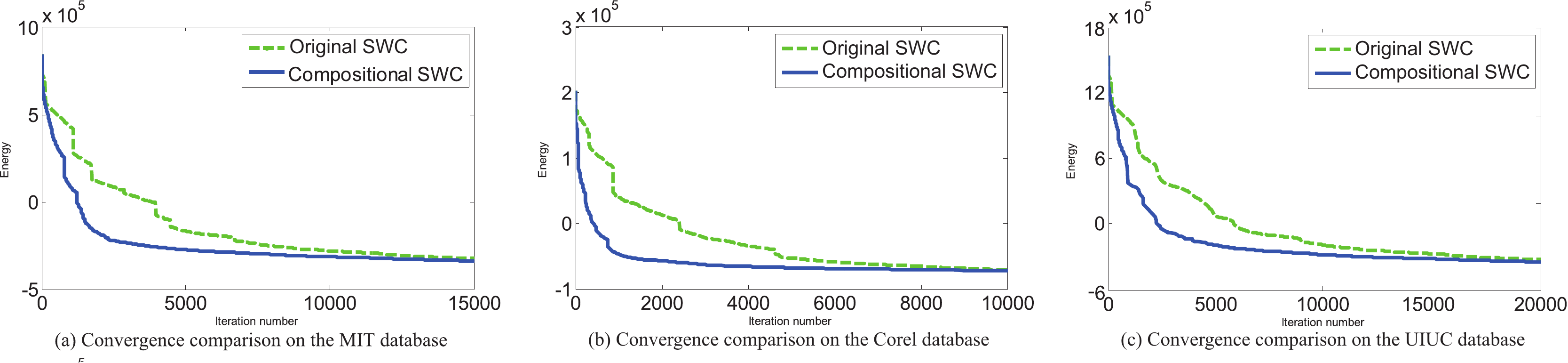}
\caption{Convergence comparisons of the CSWC algorithm and the original version. The experiments are executed on the three databases: MIT-Scene in (a), Corel in (b), and UIUC-Scene in (c). In each chart, the horizontal axis and the vertical axis, respectively, represent the iterating step and the target energy ($-\log p(S|\mathcal{D})$). The dashed (green) curves are from the original SWC algorithm and the solid (blue) curves are from the CSWC algorithm, respectively.}
\label{fig:convcomp}
\end{figure*}

\begin{figure*}[!htbp]
\centering
\includegraphics[scale=0.45]{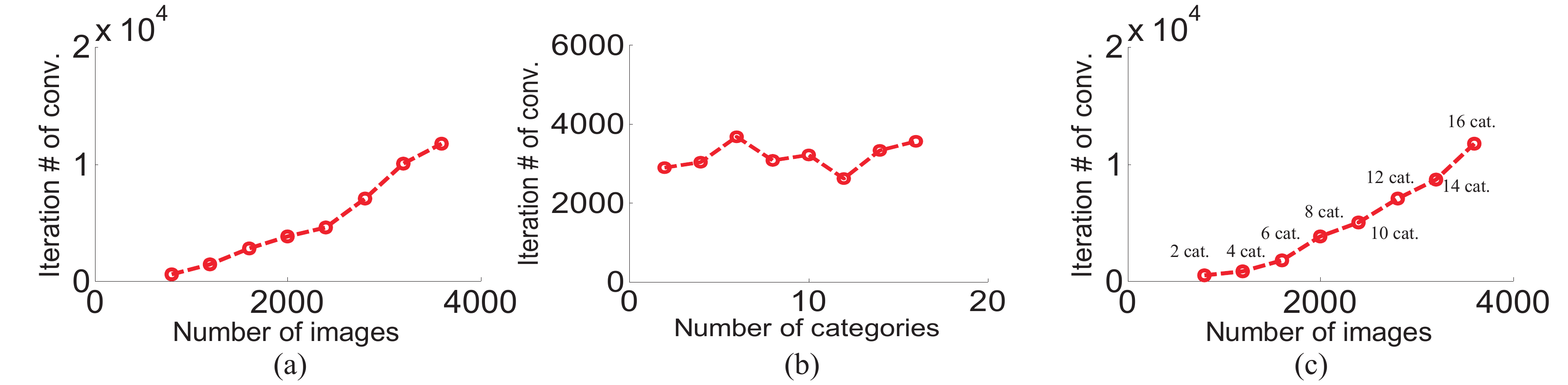}
\caption{Time complexity analysis with the increase of data scale. This analysis is performed on the mixture of MIT database and Corel database. In each figure
the vertical axis represents the speed (iteration step) of convergence; the horizontal axis in (a) represents the number of images with fixed $16$ underlying categories, in (b) the number of categories with the fixed number of images, and in (c) the number of images with various underlying categories.}
\label{fig:complexity}
\end{figure*}

For comparison, several state-of-the-art approaches are implemented based on the codes released by the original researchers, including pLSA \cite{Bosch06}, Affinity Propagation (AP)~\cite{frey07affinitypropagation} and LDA~\cite{LDA}. For the pLSA approach, we extract color SIFT descriptors to construct a dictionary of $1000$ visual words following their original implementation. For the other two approaches, i.e., AP and LDA, we use our image representations (i.e., two types of words extracted within the spatial pyramid) as the inputs of the clustering algorithms. In addition, the $k$-means clustering algorithm is adopted as the baseline, with either our representations or the gradient-based GIST features~\cite{GIST}. These methods use exactly the same experiment settings as our approach for fair evaluation, but the category number for them is manually fixed, i.e., $8$ for the MIT-Scene database, $10$ for the Corel, $15$ for the UIUC-Scene, and $23$ for the mixed dataset. The quantitative performances are reported in Table~\ref{tab:Purity} and Table \ref{tab:CondEnt} based on the two benchmark metrics, respectively. In general, our method outperforms other comparing approaches. We also evaluate our method with only one type of visual words, i.e., either ITW or HTW, so that the benefits of combining two types of features are clearly illustrated.


In our method, the clustering inference is performed simultaneously with the feature selection for category modeling. In Fig.\ref{fig:ftrselct}, we show the selected visual words of different types, i.e., ITWs and HTWs, for different categories, and the coefficients of top $40$ informative words are plotted as well. The results are very reasonable that the selected words match with the appearances of the images very well.


\subsection{Analysis}


In the following, we conduct additional empirical analysis to validate the advantages of our approach.

First, we analyze the convergence efficiency of the CSWC algorithm and compare with the original version. Fig.\ref{fig:convcomp} shows the convergence curves of the target energy, i.e., $ -\log{P(S|\mathcal{D})} $,  with the increasing iteration steps. Note that the energy goes inversely with the posterior probability. We can observe that the CSWC algorithm converges significantly faster on all the three databases.


Moreover, we analyze the computational complexity of our approach. The space complexity (i.e., computer memory) is basically related with the size of the visual word dictionary and the number of images to be categorized. Here we mainly discuss the time complexity that quantifies the amount of time taken by an algorithm conditional on the asymptotic size of the input. Using the big $\mathcal{O}$ notation, which excludes coefficients and lower order terms, the theoretic time complexity of our approach is $\mathcal{O} (MKT)$, where $M$ is the number of sampling steps, $K$ is the category number, and $T$ is the average number of features selected for each category. As we discussed in Section~\ref{sec:modellearning}, the generative model can be pursued analytically by greedy feature selection, and the feature responses on all images can be calculated off-line. In addition, only a few (i.e., $< K $) categories need to be updated in each iteration. Hence, we roughly consider the time complexity determined by the sampling steps. On the mentioned hardware, each iteration costs averagely 0.043s (MIT-Scene), 0.015s (Corel), and 0.052s (UIUC-Scene), respectively, on the three databases. In Fig.~\ref{fig:complexity}, we visualize the numbers of iteration steps on two types of data scales: the total number of images to be categorized and the underlying category number. From the results, we can observe that the steps increase in the nonexponential order, which is accordant with our analysis.

\begin{figure}[!htbp]
\centering
\includegraphics[width=3.0 in]{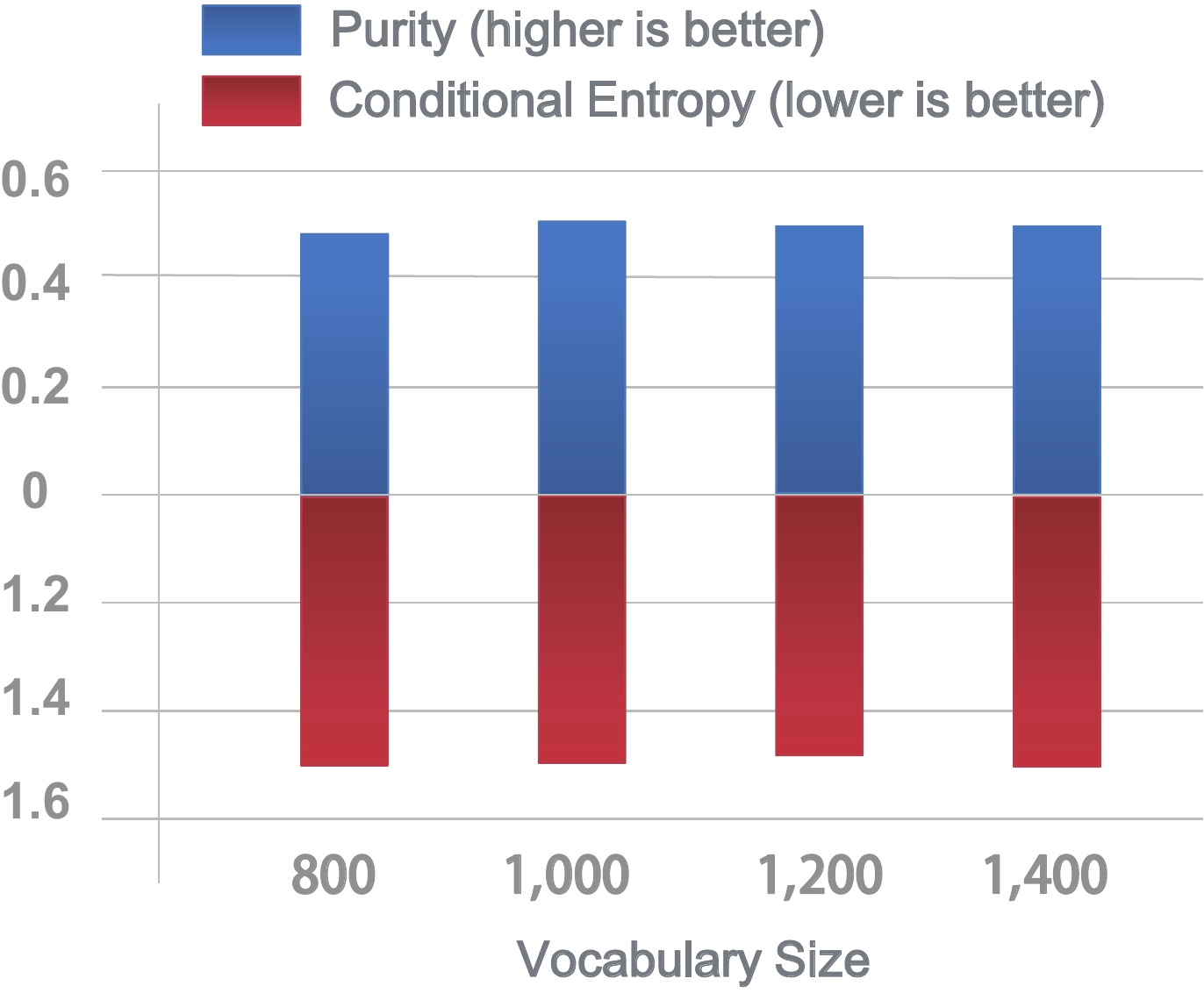}
\caption{The influence of vocabulary size. This analysis is executed on the mixed database (of $23$ categories). The upper figure and the lower figure, respectively, represent the results via {\em Purity} and {\em Conditional Entropy}. The horizontal axis represents the vocabulary size. Note that we generate equal size for the two types of words in the testings.}
\label{fig:VocSize}
\end{figure}

At last, in order to reveal how much the vocabulary size affects the results, we present an experiment in Fig.\ref{fig:VocSize}, where the categorization results are reported with different sizes of vocabulary on the mixed dataset. The conclusion can be drawn that our approach is not sensitive on the vocabulary size, as we incorporate the model learning (i.e., feature selection) with the categorization. And this property enables us to avoid elaborately tuning the size of vocabulary in practice.


\section{Conclusions}
\label{sec:conclusion}

This paper studies a general framework for automatically discovering image categories via unsupervised graph partition. Compared with the previous methods, the advantage of the proposed method is identified on several public datasets and summarized as follows. First, images are represented by two types of visual words, ITWs and HTWs, which capture image appearances from different aspects. Second, we perform feature selection simultaneously with the clustering procedure, guided by a generative model for each category. Third, we employ a stochastic sampling algorithm for efficient inference, in which the clustering number is automatically determined.

\begin{IEEEbiography}[{\includegraphics[width=1in,height=1.25in,clip,keepaspectratio]{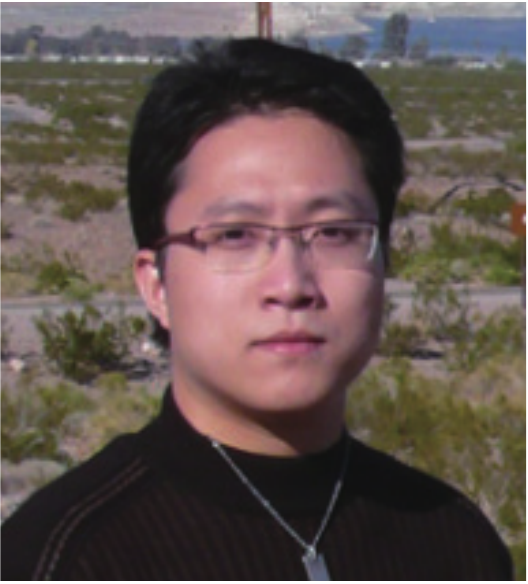}}]{Liang Lin}
Liang Lin is a full Professor with the School of Advanced Computing, Sun Yat-Sen University (SYSU), China. He received the B.S. and Ph.D. degrees from the Beijing Institute of Technology (BIT), Beijing, China, in 1999 and 2008, respectively. From 2006 to 2007, he was a joint Ph.D. student with the Department of Statistics, University of California, Los Angeles (UCLA). His Ph.D. dissertation was achieved the China National Excellent PhD Thesis Award Nomination in 2010. He was a Post-Doctoral Research Fellow with the Center for Vision, Cognition, Learning, and Art of UCLA. His research focuses on new models, algorithms and systems for intelligent processing and understanding of visual data such as images and videos. He has published more than 50 papers in top tier academic journals and conferences including Proceedings of the IEEE, T-PAMI, T-IP, T-CSVT, T-MM, Pattern Recognition, CVPR, ICCV, ECCV, ACM MM and NIPS. He was supported by several promotive programs or funds for his works, such as the �Program for New Century Excellent Talents� of Ministry of Education (China) in 2012, the �Program of Guangzhou Zhujiang Star of Science and Technology� in 2012, and the Guangdong Natural Science Funds for Distinguished Young Scholars in 2013. He received the Best Paper Runners-Up Award in ACM NPAR 2010, and Google Faculty Award in 2012. 
\end{IEEEbiography}

\begin{IEEEbiography}[{\includegraphics[width=1in,height=1.25in,clip,keepaspectratio]{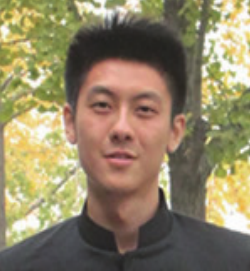}}]{Ruimao Zhang}
Ruimao Zhang received the B.E. degree in the School of Software from Sun Yat-Sen University (SYSU) in 2011. Now he is a Ph.D. candidate of Computer Science in the School of Information Science and Technology, Sun Yat-Sen University, Guangzhou, China. From 2013 to 2014, he was a visiting Ph.D. student with the Department of Computing, Hong Kong Polytechnic University (PolyU). His current research interests are computer vision, pattern recognition, machine learning and related applications. 
\end{IEEEbiography}

\begin{IEEEbiography}[{\includegraphics[width=1in,height=1.25in,clip,keepaspectratio]{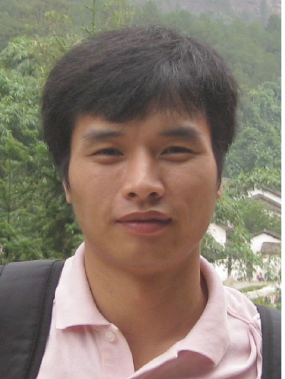}}]{Xiaohua Duan}
Xiaohua Duan received the B.B.A degree in the Department of Economic Management from Xi'an University of Posts and Telecommunications, Xi'an, China, in 2004, and the Ph.D. degree of Computer Science from Sun Yat-Sen University, Guangzhou, China, in 2012. His current research interests are image/video processing, multimedia analysis and retrieval, and computer vision. 
\end{IEEEbiography}

\end{document}